\documentclass{article}

\PassOptionsToPackage{numbers}{natbib}

    \usepackage[preprint]{neurips_2020}

\usepackage[utf8]{inputenc} 
\usepackage[T1]{fontenc}    
\usepackage{hyperref}       
\usepackage{url}            
\usepackage{booktabs}       
\usepackage{amsfonts}       
\usepackage{nicefrac}       
\usepackage{microtype}      
\usepackage{bbm}

\usepackage{graphicx} 
\usepackage{caption,subcaption}
\usepackage{multirow}

\newcommand{\eg}{\textit{e}.\textit{g}.~}

\title{Debiasing Convolutional Neural Networks via\\ Meta Orthogonalization}

\author{
    
}

\author{%
    Kurtis Evan David \\
    UT Austin \\
    \And
    Qiang Liu \\
    UT Austin \\
    \And
    Ruth Fong \\
    University of Oxford \\
}

\begin{document}

\maketitle

\begin{abstract}
  While deep learning models often achieve strong task performance, their successes are hampered by their inability to disentangle spurious correlations from causative factors, such as when they use protected attributes (\eg race, gender, etc.) to make decisions.
  In this work, we tackle the problem of debiasing convolutional neural networks (CNNs) in such instances.
  Building off of existing work on debiasing word embeddings and model interpretability, our Meta Orthogonalization method encourages the CNN representations of different concepts (\eg gender and class labels) to be orthogonal to one another in activation space while maintaining strong downstream task performance.
  Through a variety of experiments, we systematically test our method and demonstrate that it significantly mitigates model bias and is competitive against current adversarial debiasing methods.\footnote{Code available at \href{https://github.com/kurtisdavid/MetaOrthogonalization}{https://github.com/kurtisdavid/MetaOrthogonalization}}
\end{abstract}

\section{Introduction}
Recently, deep learning has been increasingly applied to a number of high-impact yet high-risk domains, like autonomous driving and medical diagnoses.
The success of deep learning has been attributed to its ability to capture complex, highly predictive patterns from large amounts of data.
However, a number of works have revealed how a variety of machine learning approaches often reflect and amplify existing human biases.

In natural language processing, \citet{debias} showed that word embeddings such as Word2Vec~\cite{word2vec} and GloVe~\cite{glove} reflected gender biases that occur naturally in most human texts and more closely associated the ``man'' concept to ``computer programmer'' and ``woman'' to ``homemaker.'' Due to the unsupervised nature of training these embeddings, it highly suggests that the corpora used to train them exhibit human and societal bias that are reflected in recorded writing \cite{Caliskan183}. As society gradually moves away from discriminating against protected classes, we may begin to have access to more unbiased data to train models. In the meantime, it is imperative that we design smarter algorithms that learn to ignore these biases while maintaining strong performance.

Of course, computer vision is no exception. Recently, \citet{pulse} has garnered much attention, showing that their facial upsampling can skew towards a Caucasian population. Much debate ensued, as societal bias can seep into the model through multiple ways, but this does not change the fact that we want our models to be independent of any harmful biases. One line of work to prevent this has focused on targeting fairness metrics primarily through adversarial learning \cite{adv}. This requires careful tuning due to the inherent min-max game, as well as the need for additional neural networks when utilizing an autoencoder in the framework \cite{laftr}. We thus pose the following: can we debias convolutional neural networks (CNNs) without adversarial learning, but instead through similar geometric arguments used for word embeddings? 

This paper empirically explores if this is possible, while using adversarial learning as a baseline to analyze any possible advantages and disadvantages. The structure of our paper is as follows. \textbf{Section 2} surveys relevant literature to our method as well as adversarial debiasing. \textbf{Section 3} frames the problem in the context of image concepts \cite{net2vec, tcav}, a method used to interpret trained CNNs, and also defines our metrics of fairness. \textbf{Section 4} then highlights our proposal, \textit{Meta Orthogonalization}; we also describe how to systematically test the various models against various situations of dataset bias. \textbf{Section 5} presents the results, including our performance on the real world COCO dataset \cite{coco}. \textbf{Section 6} summarizes and highlights our key results.



\section{Related Work}

We now survey a few recent papers closely tied to concepts in our proposal. Fairness literature has garnered much attention in order to create trustworthy models deployed in the wild. Some solutions offer post-processing the model at inference time \cite{equalityopp} to guarantee certain constraints. In a similar fashion, \citet{debias} apply a transformation to desired word embeddings to guarantee orthogonality to bias vectors in their subspace. Because this can introduce loss of downstream performance, \citet{kaneko-bollegala-2019-gender} instead augment training with a regularization term that penalizes undesired collinearity. In this paper, we follow this example, where the goal is to train a fair model without post-processing.

\subsection*{Adversarial Debiasing.} Naturally, adversarial learning \cite{gan} has seen much success in this space. Originally used to train a generator to generate realistic data from datasets, methods have been developed to utilize similar discriminator models that act as adversaries to the original model. The goal of the discriminator in this case is to be able to predict protected attributes in the dataset from information within the original model. \citet{beutelfair} apply this idea to obtain demographic parity constraints. \citet{zhang2018mitigating} extend this idea by enforcing orthogonal learning updates of the model and discriminator, to mitigate the leakage of information shared between the networks. This methodology is not restricted by dataset types; \citet{adv_fairness} apply the same framework to recidivism, and \citet{laftr} describes the same algorithm augmented with an input-level autoencoder to create fair representations that transfers across different tasks on the same dataset. Lastly, \citet{wang2018balanced} highlight the gender bias in vision systems for object detection and action recognition, and analyze the effects of adversarial learning to their introduced metric. As this paper most closely relates to our domain, we utilize it as our main baseline model comparison, and include their leakage metric for evaluation.

\subsection*{Other Relevant Work} Although adversarial learning has shown strong results, other interesting methods have been released. In particular, interpretability literature has generated methods that can easily be used in the fairness space. \citet{net2vec, tcav, ibd} learn image concept embeddings to explain fully trained models. \citet{chen2020concept} tackle the task of aligning these concepts to specific dimensions in the latent space of a network, essentially becoming an orthogonal but explainable basis. However, this is not entirely optimal, because it can be the case that certain concepts are highly correlated. This closely resembles similar ideas in our proposal; however, we only target orthogonality of image concepts w.r.t. biased information. Much work has also gone into studying inherent biases found within training datasets. \citet{cleverhans} find that major datasets contained photos that were tagged by reoccurring watermarks. Through saliency methods, they show that the neural network heavily weighted those pixels; because these watermarks do not have explicit labels, they proposed an unsupervised clustering method on the saliency maps to automatically detect various learning behaviors of the CNN, including the dependence of the logos. Similarly, \citet{bam} found that under a 1:1 co-occurrence between a pasted object and scene, the saliency maps would indicate that the CNN would pay less attention to the object, compared if only a small ratio of images had these pasted objects. 
\newpage
\section{Background}

\subsection*{Preliminary.} 
We now formalize the task explored in the rest of the paper. Let $\mathcal{D} = \{(x^{(i)}, Y^{(i)}, A^{(i)}) \}$ be our dataset of images $x^{(i)} \in \mathbb{R}^{3 \times H \times W}$, labels $Y^{(i)} \in \{ 0, 1, ... , N-1 \}$, and protected attributes $A^{(i)} \in \{0,1\}$.\footnote{In this work, we only consider binary labels for protected attributes, but recognize that this may not be representative of all possible values, e.g. gender, race, sexuality.} Our downstream task is image classification -- i.e. predicting $Y^{(i)}$ from $x^{(i)}$. To do this, we learn a CNN $f_\theta: \mathbb{R}^{3 \times H \times W} \rightarrow \mathbb{R}^N$, and given a desired layer $l$, $f^{(l)}_\theta : \mathbb{R}^{3 \times H \times W} \rightarrow \mathbb{R}^d $ will denote its intermediate representation at layer $l$. Depending on the desired downstream task, the parameters $\theta$ will be trained to minimize some loss function. In the case of image classification, this will be cross entropy.

\subsection*{Image Concepts.}
Similar to word embeddings in NLP, \textit{concepts} in images (such as colors, objects, textures) can also be represented as embedding vectors. Given $f_\theta$ suppose we have auxiliary concept labels. Given a concept $c$, we would like to learn its corresponding embedding $\beta_c$. To do this, \citet{net2vec, tcav, ibd} learn linear classifiers $g_c = \beta_c^\top z + b_c$ that operate on some intermediate representation of the network, i.e. $z_\theta(x) = f_\theta^{(l)}(x)$.\footnote{In the case that layer $l$ is convolutional, i.e. its activation is a 3D tensor, the activation is first spatially summed to produce a vector.} Concept-specific datasets are then generated and these classifiers are trained to minimize a binary log-loss for the concept. The final embedding $\beta_c$ now point in the direction to examples that are positive for concept $c$. \citet{net2vec} further observe that these concepts also have similar vector arithmetic properties as word embeddings; thus, this might suggest that we can also debias image concept embeddings through similar geometric arguments found in NLP literature \cite{debias, kaneko-bollegala-2019-gender}.

\subsection*{Fairness Measures.} To evaluate our models, we describe measures of fairness applicable to the main image classification task. The first is \textit{equality of opportunity} \cite{equalityopp}. Given a specific class $y = 0,...,N-1$ and protected attribute $A$, we would like:
\begin{equation}
    \mathrm{Pr}(\hat{Y} = y | A=0, Y=y) = \mathrm{Pr}(\hat{Y} = y | A=1, Y=y),
    \label{eq:class_eqopp}
\end{equation}

where $\hat{Y}$ represents the model prediction. Normally this measure is for binary classification, but it is extensible to the multi-class setting, as each class is mutually exclusive. To measure each model's faithfulness to this equality, we compute the absolute difference of both sides in Equation \ref{eq:class_eqopp}. We denote this as Opportunity Discrepancy for the rest of the paper.

Second, we report \textit{model leakage} -- \citet{wang2018balanced} compute this by training another classifier $h$ trained to predict the protected attribute $A$, from the final logits of the network. The resultant leakage of the model is then just its accuracy, where its ideal value is 50\%:
\begin{equation}
    \lambda(f_\theta) = \frac{1}{|\mathcal{D}|} \sum_{i=1}^{|\mathcal{D}|} \mathbbm{1}[h(f(x^{(i)})) = A^{(i)}].
    \label{eq:leakage}
\end{equation}

Third, we follow \citet{debias, kaneko-bollegala-2019-gender} by studying the geometric information found in the image concept embeddings of our trained model. Specifically, we learn a $\beta_c$ for every downstream class, in addition to two more embeddings corresponding to the positive and negative class of our protected attribute $A$. The overall bias direction is defined as $\nu = \beta_{A^{+}} - \beta_{A^{-}}$; if $\beta_c$ is the learned embedding of concept $c$, then we can measure its projection on the bias direction $\nu$. To remove the effect of magnitude, we compute the correlation between these two vectors and dub our third measure as class specific \textit{projection bias}, $\omega(c)$. If this is non-zero, then that means that the model has learned to correlate class specific information with predictive information of the protected attribute:
\begin{equation}
    \omega(c) = \frac{\beta_c^\top \nu}{||\beta_c||_2 || \nu||_2}.
    \label{eq:projb}
\end{equation}

Lastly, \citet{tcav} introduce the use of directional derivatives to explain a model, with additional post-processing. To compute the unprocessed value, we can just substitute $\beta_c$ in Equation \ref{eq:projb} with the gradient of the network for class $c$ for a specific input $x$: $\nabla_{f^{(l)}(x)}f_c(x)$. This directly connects the downstream task with bias in the model -- if these are highly correlated, then the performance of the model must be closely dependent to how it detects the bias attribute. We refer to this as \textit{sensitivity bias}. Similar arguments of sensitivity have also been used for visual interpretability in neural networks \cite{bach-plos15, montavon-pr17}.

\newcommand{\lclass}{\mathcal{L}_{\mathrm{class}}}
\newcommand{\lconcept}{\mathcal{L}_{\mathrm{concept}}}
\newcommand{\ldebias}{\mathcal{L}_{\mathrm{debias}}}

\section{Method}

\subsection*{Meta Orthogonalization.} We now propose our debiasing method. At a high level, we can decompose it into three separate losses used during training:

\begin{enumerate}
    \item \textbf{Classification Loss} ($\lclass$) -- original task loss to train the CNN, e.g. cross entropy.
    \item \textbf{Concept Loss} ($\lconcept$) -- log-loss to learn image concept vectors at a specific layer.
    \item \textbf{Debias Loss} ($\ldebias$) -- our regularization term to induce orthogonal concepts.
\end{enumerate}

\begin{figure}[hb]
    \centering 
     \includegraphics[width=0.8\linewidth]{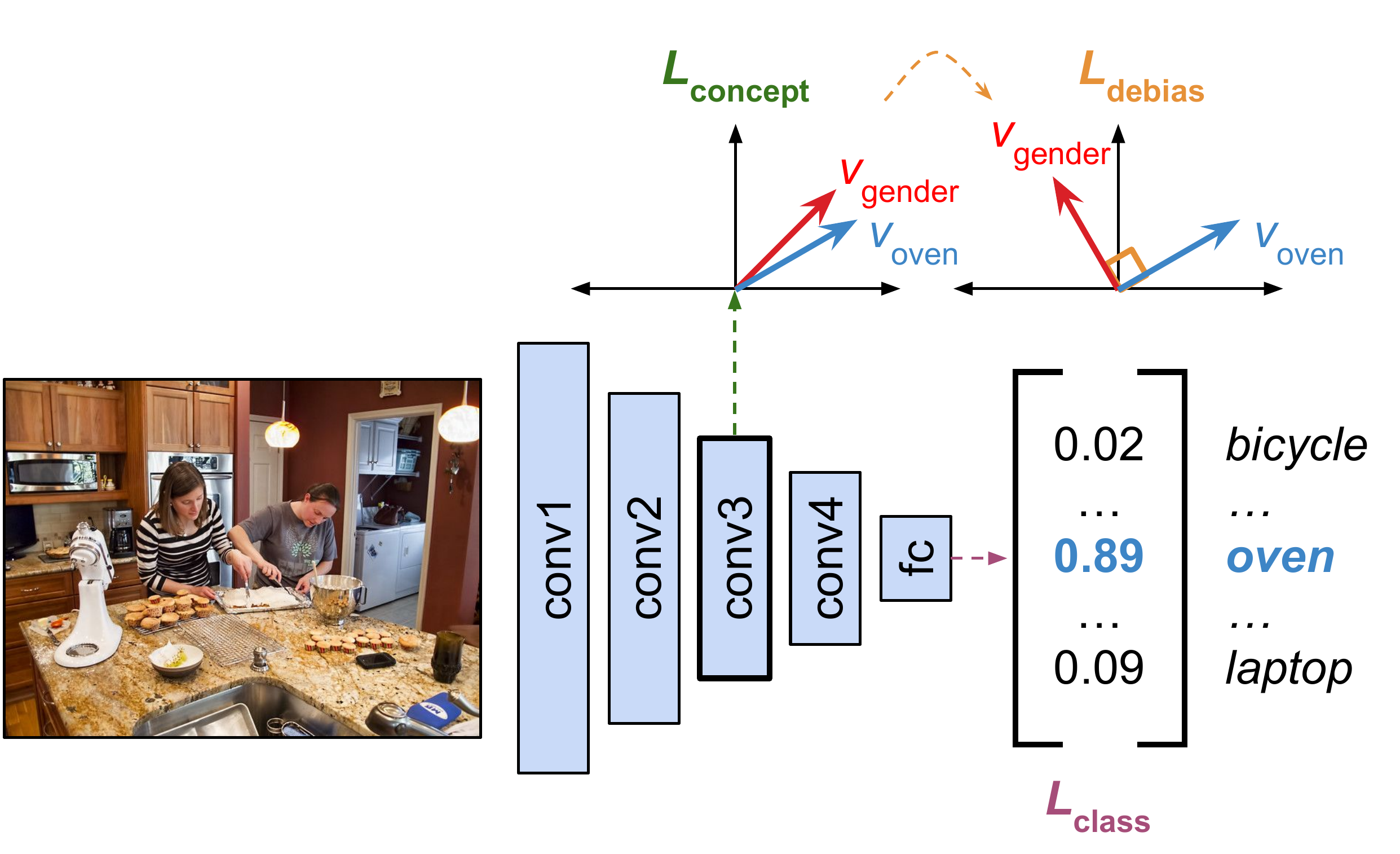}
     \label{f:diagram}
     \caption{\textbf{Method Overview.} 
     Our method debiases a CNN at an intermediate layer by encouraging the concept vector of a protected attribute and that of a downstream class to be orthogonal.
     }
\end{figure}

Both $\lclass$ and $\lconcept$ have been discussed in Section 3. The only difference is that we simultaneously learn our concept embeddings $\beta_c$, rather than at the end of training, because $\beta_c$ is directly used in our proposed regularization $\ldebias$. Following the intuition behind our \textit{projection bias} measure (Eq. \ref{eq:projb}), we initially wanted to augment training similar to \cite{kaneko-bollegala-2019-gender}, with the regularization term $\ldebias(\beta) = \sum_{c} \omega(c) ^ 2 = \sum_{c} \left(\frac{\beta_c^\top \nu}{ ||\beta_c||_2 ||\nu||_2}\right)^2$ to minimize projection bias. Although this regularizes the training of the concept embeddings, this does not affect the model parameters $\theta$ at all, since $\beta_c$ are independent parameters from the CNN. 

In order to solve this problem, we introduce the core of our algorithm: a single meta-step, inspired by similar techniques in \cite{maml, normgrad}. Assuming $\beta_c$ is learned using SGD, at every iteration, $\beta_c$ is updated to $\beta_c' = \beta_c - \alpha \nabla_{\beta_c} \lconcept(c, \theta)$. The log loss computed by $\lconcept$ is itself a function of $\theta$, and thus $\beta_c'$ is also a function of $\theta$ which can now be used to regularize the CNN:

\begin{equation}
    \ldebias(\beta') = \sum_{c} \left(\frac{\beta'^\top_c \nu}{ ||\beta'_c||_2 ||\nu||_2}\right)^2.
    \label{eq:debiasloss}
\end{equation}

In all, we summarize our method, Meta Orthogonalization, as the following minimization; when performed successfully, the goal is to end up with models whose projection bias is minimized after training:

\begin{equation}
\min_{\theta, \beta} \quad \lclass(\theta) + \sum_{c} \lconcept (c, \beta) + \gamma \ldebias(\beta^\prime).
\label{eq:final_opt}
\end{equation}

In all our tests, we use ResNet-50 \cite{resnet} as $f_\theta$. We learn image concept embeddings and apply Meta Orthogonalization at convolutional layer $\mathrm{layer3}$.\footnote{This refers to instance variables corresponding to official implementations of ResNet in PyTorch. They can be found here: https://github.com/pytorch/vision/blob/master/torchvision/models/resnet.py} To choose this layer, we observe the variance of \textit{sensitivity bias} across varying levels of controlled bias (explored in the next section). This particular layer showed the highest variance of bias, and thus we want to target this during training -- for more specifics, refer to Appendix C, Figure 1.

\subsection*{Controlling Bias.} As part of this study, we do apply our method to the COCO dataset \cite{coco}, but to extensively study the effects of our method compared to the adversarial baseline, we first create a suite of toy datasets, each with different levels of bias. To do this, we utilize the Benchmarking Attribution Methods (BAM) dataset \cite{bam}, a custom dataset specifically tailored to create object-class co-occurences. The main task is scene classification, using the MiniPlaces dataset \cite{miniplaces}. To create the co-occurrences, they extract 10 types objects from COCO to be able to crop onto the original images from MiniPlaces, at random spatial locations. With this setup, we are able to induce an artificial bias, using the presence of two object types as the protected attribute. Specifically, we choose the $\mathrm{truck}$ and $\mathrm{zebra}$ objects to represent the binary attribute $A$; we ignore the other objects for this study.

\begin{figure}[htb]
    \centering 
      \includegraphics[width=0.2\linewidth]{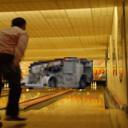} \quad
      \includegraphics[width=0.2\linewidth]{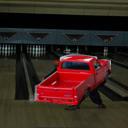} \quad
      \includegraphics[width=0.2\linewidth]{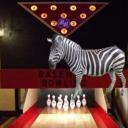} \quad
      \includegraphics[width=0.2\linewidth]{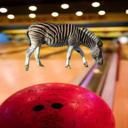} \quad
      \caption{Example images from BAM \cite{bam}, in the class ``bowling\_alley.'' \textbf{Left half:} Randomly pasted trucks. \textbf{Right half:} Randomly pasted zebras.} 
      \label{examples}
\end{figure}

We then introduce a ratio $\rho_s \in [0,1]$, representing the \% of images in a specific scene class $s$ that have a cropped truck in them; the remaining $1-\rho_s$ images of scene $s$ will have a cropped zebra. Thus, $\rho_s$ can be thought of as just the co-occurrence of the truck object with scene $s$. In the ideal case, we hope that training datasets have $\rho_s = 0.5$ for every class. To deviate from this setting, we first choose a specific \textbf{biased} subset $\mathcal{K}\subseteq [N]$ of classes that will have a different ratio $\rho_{\mathcal{K}}$; any classes not in this subset retain a ratio of 0.5. Our hope is that the model will exhibit biased behavior in this subset of images. To verify this, we do standard training the same CNN but on datasets with different $\rho_s$ for a specific class, and saw a strong linear correlation between $\rho_s$ and the class's resultant projection bias (refer to Appendix C, Figure 2). Lastly, with the introduction of subset $\mathcal{K}$, we end up with another factor to control. Specifically, we additionally study different cardinalities of $\mathcal{K}$: $\{1,3,5,7,10\}$. We randomly choose which classes will be in $\mathcal{K}$; in addition, every class in $\mathcal{K}$ will have the same ratio $\rho_{\mathcal{K}}$ of trucks/zebras.
\section{Results}

\newcommand{\minus}{\scalebox{0.75}[1.0]{$-$}}

\begin{figure}[htb]
    \centering 
      \includegraphics[width=\linewidth]{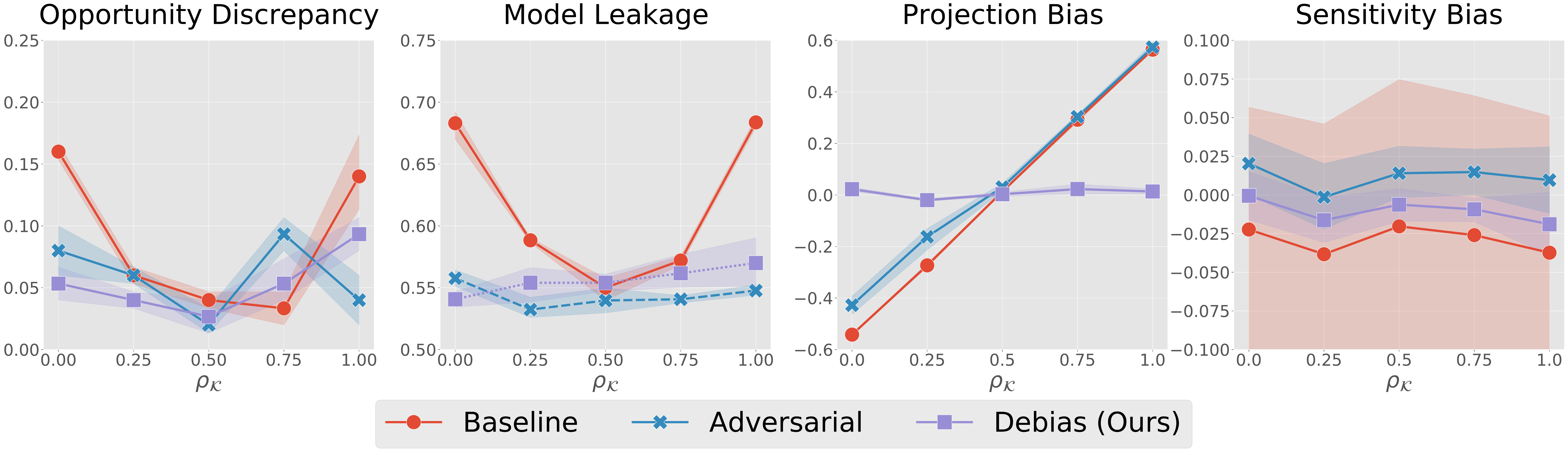}
      \caption{Measuring our four metrics on BAM, as $\rho_{\mathcal{K}}$ varies. $\mathcal{K}$ consists of a single class, ``bowling\_alley.'' Optimal shapes for Opportunity Discrepancy and Model Leakage (\textbf{left two curves}) should be flat and minimized if debiased. We see our model consistently does better than the standard training (Baseline), and is comparable to adversarial debiasing. Optimal Projection and Sensitivity Bias (\textbf{right two curves}) should at the $y=0$ line, which is closest to our model.}
      \label{fig1class}
\end{figure}

\begin{figure}[htb]
    \centering 
      \includegraphics[width=\linewidth]{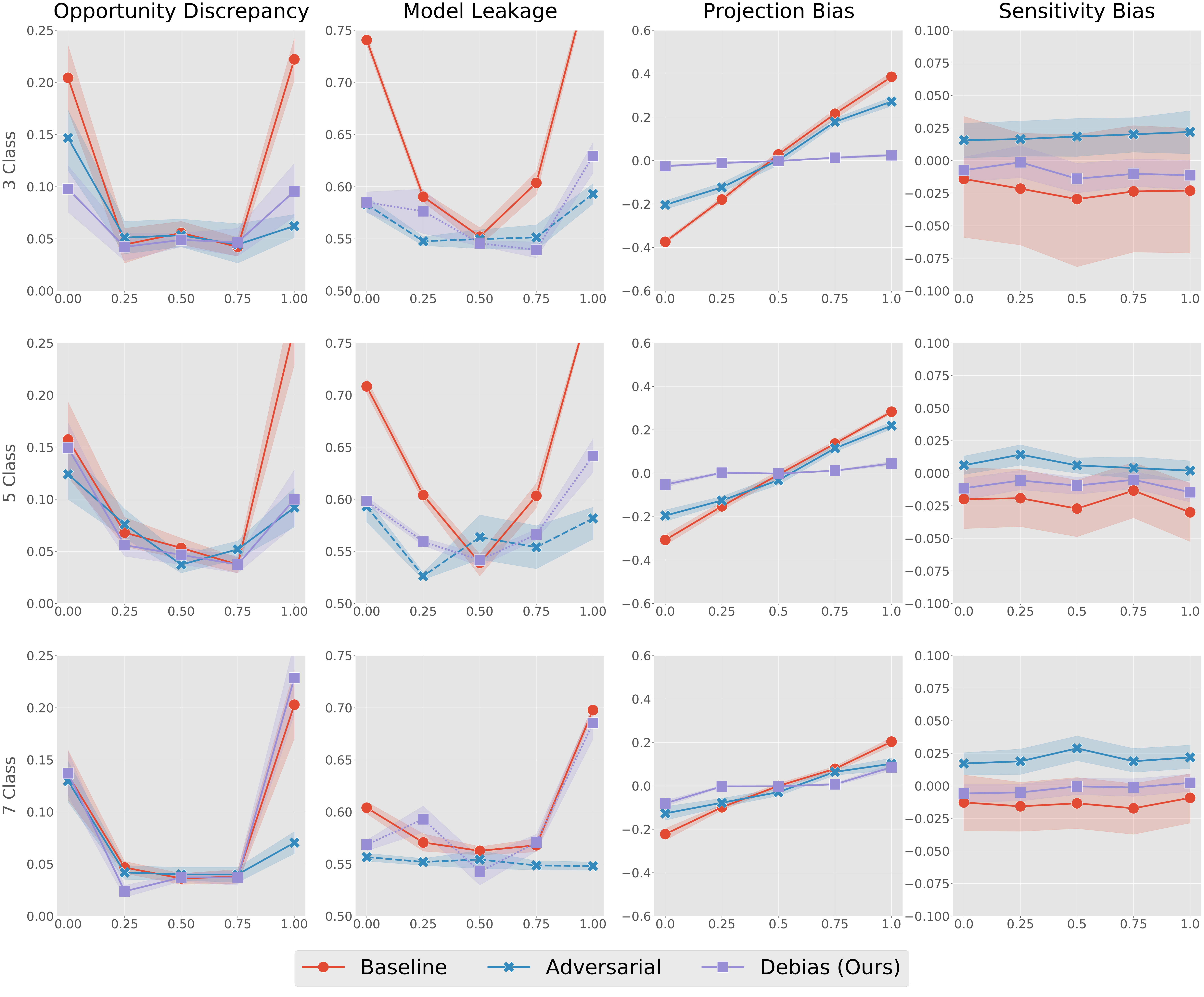}
      \caption{Analagous to Figure \ref{fig1class}, but including $|\mathcal{K}| = \{ 3, 5, 7\}$ (\textbf{Top}, \textbf{Middle}, \textbf{Bottom}, respectively). We plot the average of the class specific metrics. Our model consistently performs on par with adversarial debiasing, with added projection and sensitivity bias improvements. We do notice decreased results at the 7 Class, $\rho_{\mathcal{K}} = 1.0$ experiments, indicating that the model has not been fully debiased with our method.}
      \label{figmulticlass}
\end{figure}

\subsection*{BAM.} We now compare (1) Standard Training (Baseline), (2) Adversarial Debiasing, (3) Meta Orthogonalization Debiasing across our evaluation metrics on the various setups of BAM. Our goal is to study how varying the bias ratio $\rho_{\mathcal{K}}$ affects these models. Figure \ref{fig1class} studies this relationship when only modifying the co-occurrence ratio $\rho_{\mathcal{K}}$ in the class $\mathrm{bowling\_alley}$. In the Opportunity Discrepancy and Model Leakage curves, we see that standard training (red curve) results in clear parabolic shapes; the closer $\rho_{\mathcal{K}}$ is to 0.5, the less bias that can be observed, as every class now becomes balanced in the number of trucks and zebras pasted in the training dataset. However, once this ratio deviates from 0.5, this bias drastically increases. Therefore, optimally we would like both of these values to stay minimal and at the same level of bias found in the balanced dataset case. We find that this indeed occurs for both debiasing methods. Meta Orthogonalization (purple) remains marginally better for most ratios until 1.0, but adversarial debiasing maintains a consisitently lower model leakage. However, it is promising that our method behaves similarly to a strong debiasing method. 

In the right two curves of Figure \ref{fig1class}, we plot the projection and sensitivity biases as well. Ideally, we desire orthogonal concept vectors, therefore, the optimal measurement should be at the $y=0$ line; it is evident that Meta Orthogonalization attains the most optimal results for projection bias. This is expected, because this is exactly what our regularization term aims to do: concept class vectors learned from the trained model should be orthogonal to a learned gender direction. The projection bias of the adversarially trained model minimally deviates from standard training, indicating that any information about gender that remains in the model is still highly correlated to features useful to predict for the downstream classes. Sensitivity bias has less clear differences, but we see that ours is closer to 0, and has smaller variance compared to the two other models. We suppose this could be because the gradient can largely deviate between different inputs \cite{smilkov2017smoothgrad}, thus containing less directed information about the class and acting more similarly to randomly sampled vectors (and thus projections are already near 0) \cite{chen2020concept}.

Lastly, we include Figure \ref{figmulticlass} with our results of $|\mathcal{K}|$ = $\{3,5,7\}$. We find the case 10 to be another trivial case because the ratios within each class is now the same, but show these results separately in Appendix A, Figure 1. We find similar trends in the remaining cases, but notice weak points of our method. Notice that in the Opportunity Discrepancy column, we end up with little to no improvement in the edge $\rho_\mathcal{K}$ cases for 5 Class and 7 Class. This is similar to adversarial debiasing, but theirs remains strong at $\rho_\mathcal{K} = 1$, while ours does not. This trend is also exhibited in Model Leakage, but for the most part is similar in performance. With respect to the projection and sensitivity biases, we notice a considerable shrinking effect: the projection bias of the standard CNN decreases as the size of the biased set increases. We suppose that this occurs because the truck/zebra bias is now shared between more and more classes, and therefore the individual correlation between a class and the bias direction decreases. This is analagous to sensitivity bias, where the variance even decreases as the size of the biased set increases. Regardless, our method is still closer to 0 and thus has more desirable properties in these cases, including standard accuracy which we report in Appendix D, Table 1.

\subsection*{COCO.} To test our framework on a more real-world dataset, we opt to apply the same ideas to COCO object detection, of the images containing people. This slightly changes the framework, where instead of a single classification problem, each input has 79 predictions, one per object label. The bias attribute now refers to Male/Female images \cite{wang2018balanced}. Refer to Appendix C for details on training. Because we have separate binary classifiers, sensitivity bias is slightly modified -- we now compute the gradient of each output logit given the input has that object. As before, we average each metric across classes. We provide these values in Table \ref{coco-results}. We surprisingly find that for discrepancy, the model provided by \cite{wang2018balanced}, $\mathrm{adv @ image}$, actually increases the discrepancy value. Although \cite{wang2018balanced} provide bias interpretations through their adversarial masks in this model, this actually ends up widening the Opportunity Discrepancy, a crucial fairness statistic. On the other hand, our model indeed decreases this value, but remains within the standard deviation ranges. 

Meta Orthogonalization also provides lower leakage, even with their best model of $\mathrm{adv @ conv5}$. Even though we do not force the network to ignore gender specific features at a specific layer, we show that the logits of our model are still able to obfuscate the gender prediction by a wide margin. According to \citet{wang2018balanced}, our model almost reaches the natural leakage of this dataset $(\sim 60)$, near optimal in their case. As expected, projection bias is significantly reduced; note that the linear classifiers used for evaluation are \textit{distinct} from the ones obtained from training, to show that retraining them from scratch can also result in our desired orthogonality property. Lastly, we find that sensitivity bias in the baseline is not as prevalent, already being near 0. However, we do see our model attain marginally smaller sensitivity, in addition to a smaller variance of values, thereby mimicking the results of our BAM setup.

\begin{table}
  \caption{\textbf{COCO Results}}
  \label{coco-results}
  \centering
  \begin{tabular}{lcccc}
        \toprule
        \multicolumn{1}{c}{Model} & Discrepancy & Leakage & Projection Bias & Sensitivity Bias \\
        \midrule
        
        \multicolumn{1}{l}{$\mathrm{Baseline}$} & \multicolumn{1}{c}{$0.1314 \pm 0.0217$} & \multicolumn{1}{c}{$70.46{}^{*}$} & \multicolumn{1}{c}{$\minus0.4086 \pm 0.0243$} & \multicolumn{1}{c}{$0.0172 \pm 0.0185$} \\\cline{1-5}
        
        \multicolumn{1}{l}{$\mathrm{adv @ conv5}$} & \multicolumn{1}{c}{---} & \multicolumn{1}{c}{$64.92{}^{*}$} & \multicolumn{1}{c}{---} & \multicolumn{1}{c}{---} \\\cline{1-5}
        
        \multicolumn{1}{l}{$\mathrm{adv @ image}{}^\dagger$} & \multicolumn{1}{c}{$0.1400 \pm 0.0200$} & \multicolumn{1}{c}{$68.49{}^{*}$} & \multicolumn{1}{c}{$\minus0.4225 \pm 0.0275$} & \multicolumn{1}{c}{$0.0165 \pm 0.0141$}\\\cline{1-5}
        
        \multicolumn{1}{l}{$\mathrm{Debias (Ours)}$} & \multicolumn{1}{c}{$\mathbf{0.1245} \pm 0.0189$} & \multicolumn{1}{c}{$\mathbf{61.07}^{\phantom{*}}$} & \multicolumn{1}{c}{$\phantom{\minus}\textbf{0.0003} \pm 0.0416$} & \multicolumn{1}{c}{$\mathbf{0.0066} \pm 0.0067$} \\
        \bottomrule
    \end{tabular}
    \caption*{As before, lower is better for Discrepancy and Leakage, and closer to 0 is better for Projection and Sensitivity. We find our debiasing algorithm results in more favorable results across the board. We also provide standard errors for Discrepancy, and standard deviations for Projection Bias and Sensitivity Bias, across COCO object classes. \ $\mathbf{*}$ denotes obtained from \cite{wang2018balanced}, Table 4.  $\mathbf{\dagger}$ denotes results using model provided by \cite{wang2018balanced}.}
\end{table}

\section{Discussion}

We now highlight a few key points we notice in our results. First, Meta Orthoganalization almost always performs comparably with Adversarial Debiasing across our evaluation metrics. The main situation that it fails is when $|\mathcal{K}|=7$ when studying BAM. We think more careful hyperparameter tuning may fix this, but were unable to do so that worked across all situations in Figures 3 and 4. This is important, in that real datasets may not have an expected measure of bias as we did when controlling $\rho_{\mathcal{K}}$. 

Another finding is that in terms of the fairness metrics, opportunity discrepancy and model leakage, our method is able to obtain ideal results, even though we do not explicitly encode for this in the loss. Our regularization term does not enforce as strong of a constraint as an adversarial loss (i.e. removing necessary information for predicting the protected class), but can still be fair according to definitions in literature. In addition, our method is still able to retain predictive information for all concepts and protected attributes, but by applying the meta-loss, disentangles this information from the downstream classification task. We leave possible theoretical connections with our method and these mathematical definitions to future work. 

Lastly, through the varying parameters on $\mathcal{K}$, we provide more evidence that the more co-occuring sets of objects are, CNNs tend to ignore this information \cite{bam}. Specifically, when we increase the size of the biased set, the breadth of bias shrinks across varying ratios. To see the effect on the unbiased set, refer to Appendix A, Figure 1. We see more increased bias in the ``unbiased'' set, even though the have equalized ratio of the protected attribute. In addition, whenever $\mathcal{K}$ encompasses all labels in the dataset, nearly zero bias is seen across the evaluation metrics.

\subsection*{Future Work.} As mentioned, it is still open as to the mathematical connection between orthogonal concept directions and specific fairness definitions; as of this point we only have empirical evidence. An interesting idea that could work is a slight modification to the Debias Loss (Equation \ref{eq:debiasloss}). Instead of enforcing class concepts to be completely orthogonal to the bias direction, we can have a weakened version, based off of a threshold $t$. This may be useful by having different thresholds per class, since in real datasets, every class will be affected in different ways, rather than sharing a uniform biased ratio $\rho$. Lastly, it would be interesting to see how this can affect more complex downstream tasks that would utilize the debiased CNNs from our method, rather than a simple prediction. For example, \citet{women_snowboard} aim to debias image captions. Meta Orthogonalization can still be applied, because the loss does not take into account anything about the downstream task; it would be interesting to see if it can also avoid biased image captions.

\section{Conclusion}

In this work, we propose \textit{Meta Orthogonalization} as a way to debias convolutional neural networks by pushing image concepts to be orthogonal to a learned bias direction. In addition, we provide a methodology to simulate various levels of bias and situations in a dataset, through the use of BAM \cite{bam}, a recent dataset of object co-occurrences. We show that across various fairness metrics, our method performs comparably, if not better, to adversarial debiasing, even on COCO \cite{coco}. Through extensive analysis on various trends of bias, we have shown promising empirical results for our method, indicating that there may be a strong connection to bias learned in CNNs and geometric properties of their activation spaces.
\newpage

\section*{Broader Impact}

Systemic bias is a pervasive issue today, and as more complex models are used by governing bodies, it is imperative that we can provide methods that will mitigate the propagation of harmful biases. Currently, adversarial debiasing has portrayed its strengths in many applications, but we have shown that other methods may be needed, especially for image tasks. We show that enforcing concept orthogonality, a natural objective, in CNNs is a viable and easily interpretable alternative to doing so. This can open up increased possibility for researchers to explore other ideas integrating fairness and interpretability together. It must be said, however, that practictioners should take caution in utilizing this framework, or any other debiasing algorithm. Our datasets in particular have been highly controlled, and do not fully reflect bias that may occur in practical applications. It should always be important to test simple fairness definitions before fully deploying debiased models.

\bibliographystyle{plainnat}
\bibliography{ref}

\begin{thebibliography}{29}
\providecommand{\natexlab}[1]{#1}
\providecommand{\url}[1]{\texttt{#1}}
\expandafter\ifx\csname urlstyle\endcsname\relax
  \providecommand{\doi}[1]{doi: #1}\else
  \providecommand{\doi}{doi: \begingroup \urlstyle{rm}\Url}\fi

\bibitem[Bach et~al.(2015)Bach, Binder, Montavon, Klauschen, M{\"u}ller, and
  Samek]{bach-plos15}
Sebastian Bach, Alexander Binder, Gr{\'e}goire Montavon, Frederick Klauschen,
  Klaus-Robert M{\"u}ller, and Wojciech Samek.
\newblock On pixel-wise explanations for non-linear classifier decisions by
  layer-wise relevance propagation.
\newblock \emph{PLoS ONE}, 10, 07 2015.

\bibitem[Beutel et~al.(2017)Beutel, Chen, Zhao, and Chi]{beutelfair}
Alex Beutel, Jilin Chen, Zhe Zhao, and Ed~H. Chi.
\newblock Data decisions and theoretical implications when adversarially
  learning fair representations.
\newblock \emph{CoRR}, abs/1707.00075, 2017.

\bibitem[Bolukbasi et~al.(2016)Bolukbasi, Chang, Zou, Saligrama, and
  Kalai]{debias}
Tolga Bolukbasi, Kai-Wei Chang, James~Y Zou, Venkatesh Saligrama, and Adam~T
  Kalai.
\newblock Man is to computer programmer as woman is to homemaker? debiasing
  word embeddings.
\newblock In D.~D. Lee, M.~Sugiyama, U.~V. Luxburg, I.~Guyon, and R.~Garnett,
  editors, \emph{Advances in Neural Information Processing Systems 29}, pages
  4349--4357. Curran Associates, Inc., 2016.

\bibitem[Caliskan et~al.(2017)Caliskan, Bryson, and Narayanan]{Caliskan183}
Aylin Caliskan, Joanna~J. Bryson, and Arvind Narayanan.
\newblock Semantics derived automatically from language corpora contain
  human-like biases.
\newblock \emph{Science}, 356\penalty0 (6334):\penalty0 183--186, 2017.
\newblock ISSN 0036-8075.
\newblock \doi{10.1126/science.aal4230}.
\newblock URL \url{https://science.sciencemag.org/content/356/6334/183}.

\bibitem[Chen et~al.(2020)Chen, Bei, and Rudin]{chen2020concept}
Zhi Chen, Yijie Bei, and Cynthia Rudin.
\newblock Concept whitening for interpretable image recognition.
\newblock In \emph{arXiv}, 2020.

\bibitem[Finn et~al.(2017)Finn, Abbeel, and Levine]{maml}
Chelsea Finn, Pieter Abbeel, and Sergey Levine.
\newblock Model-agnostic meta-learning for fast adaptation of deep networks.
\newblock In \emph{Proceedings of the 34th International Conference on Machine
  Learning - Volume 70}, ICML’17, page 1126–1135. JMLR.org, 2017.

\bibitem[Fong and Vedaldi(2018)]{net2vec}
Ruth Fong and Andrea Vedaldi.
\newblock Net2vec: Quantifying and explaining how concepts are encoded by
  filters in deep neural networks.
\newblock In \emph{Proceedings of the IEEE Conference on Computer Vision and
  Pattern Recognition (CVPR)}, June 2018.

\bibitem[Ganin et~al.(2016)Ganin, Ustinova, Ajakan, Germain, Larochelle,
  Laviolette, Marchand, and Lempitsky]{adv}
Yaroslav Ganin, Evgeniya Ustinova, Hana Ajakan, Pascal Germain, Hugo
  Larochelle, François Laviolette, Mario Marchand, and Victor~S. Lempitsky.
\newblock Domain-adversarial training of neural networks.
\newblock \emph{J. Mach. Learn. Res.}, 17:\penalty0 59:1--59:35, 2016.

\bibitem[Goodfellow et~al.(2014)Goodfellow, Pouget-Abadie, Mirza, Xu,
  Warde-Farley, Ozair, Courville, and Bengio]{gan}
Ian Goodfellow, Jean Pouget-Abadie, Mehdi Mirza, Bing Xu, David Warde-Farley,
  Sherjil Ozair, Aaron Courville, and Yoshua Bengio.
\newblock Generative adversarial nets.
\newblock In Z.~Ghahramani, M.~Welling, C.~Cortes, N.~D. Lawrence, and K.~Q.
  Weinberger, editors, \emph{Advances in Neural Information Processing Systems
  27}, pages 2672--2680. Curran Associates, Inc., 2014.

\bibitem[Hardt et~al.(2016)Hardt, Price, Price, and Srebro]{equalityopp}
Moritz Hardt, Eric Price, Eric Price, and Nati Srebro.
\newblock Equality of opportunity in supervised learning.
\newblock In D.~D. Lee, M.~Sugiyama, U.~V. Luxburg, I.~Guyon, and R.~Garnett,
  editors, \emph{Advances in Neural Information Processing Systems 29}, pages
  3315--3323. Curran Associates, Inc., 2016.

\bibitem[{He} et~al.(2016){He}, {Zhang}, {Ren}, and {Sun}]{resnet}
K.~{He}, X.~{Zhang}, S.~{Ren}, and J.~{Sun}.
\newblock Deep residual learning for image recognition.
\newblock In \emph{2016 IEEE Conference on Computer Vision and Pattern
  Recognition (CVPR)}, pages 770--778, 2016.

\bibitem[Hendricks et~al.(2018)Hendricks, Burns, Saenko, Darrell, and
  Rohrbach]{women_snowboard}
Lisa~Anne Hendricks, Kaylee Burns, Kate Saenko, Trevor Darrell, and Anna
  Rohrbach.
\newblock Women also snowboard: Overcoming bias in captioning models.
\newblock In Vittorio Ferrari, Martial Hebert, Cristian Sminchisescu, and Yair
  Weiss, editors, \emph{Computer Vision -- ECCV 2018}, pages 793--811, Cham,
  2018. Springer International Publishing.
\newblock ISBN 978-3-030-01219-9.

\bibitem[Kaneko and Bollegala(2019)]{kaneko-bollegala-2019-gender}
Masahiro Kaneko and Danushka Bollegala.
\newblock Gender-preserving debiasing for pre-trained word embeddings.
\newblock In \emph{Proceedings of the 57th Annual Meeting of the Association
  for Computational Linguistics}, Florence, Italy, July 2019. Association for
  Computational Linguistics.

\bibitem[Kim et~al.(2018)Kim, Wattenberg, Gilmer, Cai, Wexler, Vi{\'e}gas, and
  Sayres]{tcav}
Been Kim, Martin Wattenberg, Justin Gilmer, Carrie~J. Cai, James Wexler,
  Fernanda~B. Vi{\'e}gas, and Rory Sayres.
\newblock Interpretability beyond feature attribution: Quantitative testing
  with concept activation vectors (tcav).
\newblock In \emph{ICML}, 2018.

\bibitem[Lapuschkin et~al.(2019)Lapuschkin, Wäldchen, Binder, Montavon, Samek,
  and Müller]{cleverhans}
Sebastian Lapuschkin, Stephan Wäldchen, Alexander Binder, Grégoire Montavon,
  Wojciech Samek, and Klaus-Robert Müller.
\newblock Unmasking clever hans predictors and assessing what machines really
  learn.
\newblock \emph{Nature Communications}, 10\penalty0 (1), Mar 2019.
\newblock ISSN 2041-1723.
\newblock \doi{10.1038/s41467-019-08987-4}.
\newblock URL \url{http://dx.doi.org/10.1038/s41467-019-08987-4}.

\bibitem[Lin et~al.(2014)Lin, Maire, Belongie, Hays, Perona, Ramanan,
  Doll{\'a}r, and Zitnick]{coco}
Tsung-Yi Lin, Michael Maire, Serge Belongie, James Hays, Pietro Perona, Deva
  Ramanan, Piotr Doll{\'a}r, and C.~Lawrence Zitnick.
\newblock Microsoft coco: Common objects in context.
\newblock In David Fleet, Tomas Pajdla, Bernt Schiele, and Tinne Tuytelaars,
  editors, \emph{Computer Vision -- ECCV 2014}, pages 740--755, Cham, 2014.
  Springer International Publishing.
\newblock ISBN 978-3-319-10602-1.

\bibitem[Madras et~al.(2018)Madras, Creager, Pitassi, and Zemel]{laftr}
David Madras, Elliot Creager, Toniann Pitassi, and Richard Zemel.
\newblock Learning adversarially fair and transferable representations.
\newblock In Jennifer Dy and Andreas Krause, editors, \emph{Proceedings of the
  35th International Conference on Machine Learning}, volume~80 of
  \emph{Proceedings of Machine Learning Research}, pages 3384--3393,
  Stockholmsmässan, Stockholm Sweden, 10--15 Jul 2018. PMLR.

\bibitem[Menon et~al.(2020)Menon, Damian, Hu, Ravi, and Rudin]{pulse}
Sachit Menon, Alexandru Damian, Shijia Hu, Nikhil Ravi, and Cynthia Rudin.
\newblock Pulse: Self-supervised photo upsampling via latent space exploration
  of generative models.
\newblock In \emph{Proceedings of the IEEE/CVF Conference on Computer Vision
  and Pattern Recognition (CVPR)}, June 2020.

\bibitem[Mikolov et~al.(2013)Mikolov, Sutskever, Chen, Corrado, and
  Dean]{word2vec}
Tomas Mikolov, Ilya Sutskever, Kai Chen, Greg~S Corrado, and Jeff Dean.
\newblock Distributed representations of words and phrases and their
  compositionality.
\newblock In C.~J.~C. Burges, L.~Bottou, M.~Welling, Z.~Ghahramani, and K.~Q.
  Weinberger, editors, \emph{Advances in Neural Information Processing Systems
  26}, pages 3111--3119. Curran Associates, Inc., 2013.

\bibitem[Montavon et~al.(2017)Montavon, Bach, Binder, Samek, and
  M{\"u}ller]{montavon-pr17}
Gr{\'e}goire Montavon, Sebastian Bach, Alexander Binder, Wojciech Samek, and
  Klaus-Robert M{\"u}ller.
\newblock Explaining nonlinear classification decisions with deep taylor
  decomposition.
\newblock \emph{Pattern Recognition}, 65:\penalty0 211--222, 2017.
\newblock \doi{10.1016/j.patcog.2016.11.008}.

\bibitem[Pennington et~al.(2014)Pennington, Socher, and Manning]{glove}
Jeffrey Pennington, Richard Socher, and Christopher~D. Manning.
\newblock Glove: Global vectors for word representation.
\newblock In \emph{Empirical Methods in Natural Language Processing (EMNLP)},
  pages 1532--1543, 2014.
\newblock URL \url{http://www.aclweb.org/anthology/D14-1162}.

\bibitem[Rebuffi et~al.(2020)Rebuffi, Fong, Ji, and Vedaldi]{normgrad}
Sylvestre-Alvise Rebuffi, Ruth~C. Fong, Xu~Ji, and Andrea Vedaldi.
\newblock There and back again: Revisiting backpropagation saliency methods.
\newblock In \emph{Proceedings of the {IEEE} Conference on Computer Vision and
  Pattern Recognition ({CVPR})}, 2020.

\bibitem[Smilkov et~al.(2017)Smilkov, Thorat, Kim, Viégas, and
  Wattenberg]{smilkov2017smoothgrad}
Daniel Smilkov, Nikhil Thorat, Been Kim, Fernanda Viégas, and Martin
  Wattenberg.
\newblock Smoothgrad: removing noise by adding noise, 2017.

\bibitem[Wadsworth et~al.(2018)Wadsworth, Vera, and Piech]{adv_fairness}
Christina Wadsworth, Francesca Vera, and Chris Piech.
\newblock Achieving fairness through adversarial learning: an application to
  recidivism prediction.
\newblock \emph{CoRR}, abs/1807.00199, 2018.

\bibitem[Wang et~al.(2019)Wang, Zhao, Yatskar, Chang, and
  Ordonez]{wang2018balanced}
Tianlu Wang, Jieyu Zhao, Mark Yatskar, Kai-Wei Chang, and Vicente Ordonez.
\newblock Balanced datasets are not enough: Estimating and mitigating gender
  bias in deep image representations.
\newblock \emph{2019 IEEE/CVF International Conference on Computer Vision
  (ICCV)}, pages 5309--5318, 2019.

\bibitem[Yang and Kim(2019)]{bam}
Mengjiao Yang and Been Kim.
\newblock {Benchmarking Attribution Methods with Relative Feature Importance}.
\newblock \emph{CoRR}, abs/1907.09701, 2019.

\bibitem[Zhang et~al.(2018)Zhang, Lemoine, and Mitchell]{zhang2018mitigating}
Brian~Hu Zhang, Blake Lemoine, and Margaret Mitchell.
\newblock Mitigating unwanted biases with adversarial learning.
\newblock In \emph{Proceedings of the 2018 AAAI/ACM Conference on AI, Ethics,
  and Society}, AIES '18, page 335–340, New York, NY, USA, 2018. Association
  for Computing Machinery.
\newblock ISBN 9781450360128.
\newblock \doi{10.1145/3278721.3278779}.
\newblock URL \url{https://doi.org/10.1145/3278721.3278779}.

\bibitem[Zhou et~al.(2017)Zhou, Lapedriza, Khosla, Oliva, and
  Torralba]{miniplaces}
Bolei Zhou, Agata Lapedriza, Aditya Khosla, Aude Oliva, and Antonio Torralba.
\newblock Places: A 10 million image database for scene recognition.
\newblock \emph{IEEE Transactions on Pattern Analysis and Machine
  Intelligence}, 2017.

\bibitem[Zhou et~al.(2018)Zhou, Sun, Bau, and Torralba]{ibd}
Bolei Zhou, Yiyou Sun, David Bau, and Antonio Torralba.
\newblock Interpretable basis decomposition for visual explanation.
\newblock In \emph{Proceedings of the European Conference on Computer Vision
  (ECCV)}, September 2018.

\end{thebibliography}

\end{document}